# Exploring Transformers in Emotion Recognition: a comparison of BERT, DistillBERT, RoBERTa, XLNet and ELECTRA

Diogo Cortiz
Brazilian Network Information Center (NIC.br)
Pontifical Catholic University of São Paulo (PUC-SP)
dcortiz@pucsp.br

**Abstract**

This paper investigates how Natural Language Understanding (NLU) could be applied in Emotion Recognition, a specific task in affective computing. We finetuned different transformers language models (BERT, DistilBERT, RoBERTa, XLNet, and ELECTRA) using a fine-grained emotion dataset and evaluating them in terms of performance (f1-score) and time to complete.

## 1 Introduction

One of the main challenges of affective computing involving the Natural Language Understanding (NLU) area is recognizing emotions. Some projects seek to understand emotions through facial expressions, body movement, and blood pressure. In this paper, we will delimit the scope for recognizing emotional expressions in texts.

We have seen different architectures emerging to deal with challenges in NLU tasks, such as RNNs, LSTM, BiLSTM. More recently, the Transformers architecture (Vaswani et al., 2017) has shown to be an appropriate alternative for better results in NLU tasks.

Transformers' language models have shown better language understanding abilities to achieve state-of-the-art results in many different tasks. Our goal is to investigate the performance of different transformer language models for Emotion Recognition when using a dataset of fine-grained emotions.

## 2 Related work

This section reviews five articles that propose different approaches, methods, or techniques for recognizing emotions in texts. One study (Polignano et al., 2019) proposed an approach based on deep neural network. In this case, the researchers demonstrated that an architecture composed of BiLSTM, CNN, and self-attention demonstrated promising results in different datasets. They also took the opportunity to compare three pre-trained word-embeddings (Google word embeddings, GloVe, and FastText) for word encoding.

The authors argued the model is based on the synergy of two deep learning approaches for classification: long-short-term memory networks (LSTM) in their bi-directional variation (BiLSTM) and the convolutional neural networks (CNN) mediated by approach max pooling. However, the architectural difference includes a self-attention layer after BiLSTM so that the model captures distant relationships between words, with different weights according to their contribution to the classification task.

For embedding layers, the authors argue that training a word embedding directly on the application domain's "training data" can impact the model's generalization ability. If a new sentence is used as an entry for classification, words are missing leaving the task difficult to complete successfully.

To overcome this challenge, the authors used transfer learning practice, using word embeddings already pre-trained in different domains. With that, they could reduce the computational cost while covering a wider range of terms regardless of domains. They applied GoogleEmb, GloVe, and FastText in different experiments and compared them. FastText has shown very significance for highlighting the better performances on all the three datasets evaluated.

Authors used three datasets: **ISEAR** (Scherer and Wallbott, 1994), which contains annotation for seven emotions (joy, fear, anger, sadness, disgust, shame, and guilt); **SemEval 2018 task 1**, which contains annotations only for joy, fear, anger, and sadness; and **SemEval 2019 task 3**, which contains three specifically annotated emotions: happy, sad, angry and an 'other' class that includes all others not annotated possible emotions.

| Model | Embeddings | Macro-F1 |
|---|---|---|
| SVM | - | 0.55 |
| Random Forest | - | 0.49 |
| BiLSTM+CNN+Self-Attention | GoobleEmb | 0.62 |
| BiLSTM+CNN+Self-Attention | GloVE | 0.62 |
| BiLSTM+CNN+Self-Attention | FastText | 0.63 |

Table 1: Comparison of BiLSTM+CNN+Self-Attention models for ISEAR

The model's results combining BiLSTM, CNN, and Self-attention demonstrated the effectiveness of this approach compared to baseline models and some other deep neural approaches. The authors also point out that the choice of encoding the textual input through word embedding can influence the behavior and the final result of the model. The results showed that FastText could bring better results to the task of classifying emotional sentences.

Batbaatar et al. (2019) argued that several previous research pieces adopted word embedding vectors that could represent rich semantic/syntactic information but had difficulty identifying emotional relationships between words. Some word embeddings with emotional information have been proposed, but it also requires semantic and syntactic information. Those models' approaches cannot encode and learn semantic and emotional relationships efficiently in a short text.

To overcome this challenge, the authors proposed a new neural network architecture (called semantic-emotion neural network - SENN) that combines two sub-networks to capture semantic and emotional information. The first network consists of a BiLSTM to capture semantic information and map it into semantic-sentence space; the second network is a CNN to capture emotional information and map it to an emotion-sentence space.

According to the authors, CNN is supposedly suitable for extracting position invariant features (emotion information), while BiLSTM is indicated to deal with a semantic sequence in a sentence (contextual information). The combination of the final representations can be helpful for the recognition of emotions. The study used different pre-trained general word embeddings (Word2Vec, GloVE, and FastText) and an emotion-enriched word embedding (EWE).

The authors explored different available datasets (from distinct domains: social media, fairy tale, etc.). We highlight the **ISEAR** dataset for this literature review because it is the common dataset used by other papers we discuss. It is important to note that the authors made some kind of transformation in the data that was not very clear in the paper. The original dataset contains 7,666 annotated sentences, but the authors used only 5,241 sentences in the SENN model experiment.

The results showed that the SENN model performed better than baseline models and other state-of-the-art deep learning models to classify datasets based on Paul Ekman's six categories of emotions (using different datasets). Specifically, for the ISEAR dataset, the performance of the SENN model is highlighted above:

| Model | Embeddings | Macro-F1 |
|---|---|---|
| SENN | Word2Vec + EWE | 0.737 |
| SENN | GloVe + EWE | 0.746 |
| SENN | FastText + EWE | 0.745 |

Table 2: Comparison SENN models for ISEAR

Even though GloVe combined with EWE performed slightly better for ISEAR, the authors observed that among all variants of the SENN model, the one that used FastText performed better than Word2Vec and Glove. The authors argue that a possible explanation is that FastText word embeddings capture the meaning of smaller words and allow embeddings to understand suffixes and prefixes, facilitating the processing of rare words and out-of-vocabulary. Based on the results, we may discuss that using a specific embedding for emotions can improve emotional sentence classification tasks.

In a recent publication (Adoma et al., 2020), there was a comparative analysis of different Transformer-based models (BERT, RoBERTa, DistilBert, and XLNet) in an emotion recognition task, using the ISEAR dataset Authors argued the difficulty of identifying the most appropriate embeddings for extracting long sequence relationships and capturing lightning information in a specific scenario (such as emotion recognition), can be overcome with the Transformers approach and language models based on transformers.

This position appears to agree with the results of the previously reviewed papers. They noticed that embeddings directly influence the model's behavior and that task-related embeddings (as was the case with emotion-enriched word embedding - EWE) can bring better results.

The purpose of the paper is to analyze the effectiveness of the BERT, RoBERTa, DistilBERT, and XLNET models in recognizing emotions using the ISEAR dataset. Each model's experiments and results are discussed comparatively according to accuracy, precision, recall, and F1-score for each of the classes of emotions in the dataset.

Unlike the other reviewed articles, the authors detailed the data pre-processing. They removed records in which there was a column of emotions, but the sentence was missing. Thus, the dataset used in the experiment reduced in size, from 7666 sentences to 7589 sentences. The authors also stated that they removed special characters, double spaces, tags, and irregular expressions because they could negatively affect the models' performance. Stop words have also been removed. The dataset was divided into 80% of the data for training and 20% for testing purposes.

All models were trained using the same hyperparameters. The results showed RoBERTa as the best model (accuracy of 0.743), followed by XLNet (accuracy of 0.729). BERT achieved the third position (accuracy of 0.700) while DistillBERT holds the last position (accuracy of 0.669). In Table 3, we present the Macro-F1 for each class.

| Model | Macro-F1 |
|---|---|
| BERT | 0.702 |
| RoBERTa | 0.742 |
| DistilBERT | 0.693 |
| XLNet | 0.731 |

Table 3: Comparison of Transformers models for ISEAR

The authors concluded that RoBERTa pre-trained model outperforms the other pre-trained models. They also discuss that even though DistilBERT was the least accurate model, it was the fastest model while XLNet was the slowest. The decreasing order of computational resources is given as XLNet, BERT, RoBERTa, DistilBERT.

Finally, the authors conclude that pre-trained models based on transformers prove to be effective in detecting emotions in texts, with RoBERTa presenting the best metrics for accuracy and macro-f1. This work is important to show the potential use of pre-trained models in specific scenarios (as in recognizing emotions), just finetuning the model.

The papers reviewed so far reflect a common characteristic in NLP projects that deal with emotions: the assumption that emotions are structured in just six basic emotions. Most datasets available (such as the ISEAR dataset) are structured based on this theoretical position ranging from 5 to 6 classes of emotions (on average).

However, there is a debate in affective science, neuroscience, psychology, and philosophy about what an emotion is and how many there are. We can divide the positions into two approaches: basic emotion theory (BET) and constructivism. BET assumes that only a few discrete emotions are separated by clear boundaries and that categories are universal. On the other hand, constructivism argues that emotions are constructions based on valence and arousal, that people put different interpretations for affection from an individual perspective.

There is an alternative perspective: semantic space theory (Cowen and Keltner, 2021). The authors describe this theory as a computational approach that explores broad natural stimuli and open-ended statistical techniques to capture systematic variation in behaviors related to emotions.

The authors argued that there are more than 25 distinct emotional experience varieties with different backgrounds and expressions. These emotions are highly dimensional, categorical, and can be blended. Semantic space theory is a promising proposal and is arousing interest in research in affective science. In this sense, it is also essential to investigate their possible interaction in emotion recognition projects using NLU approaches.

Recently, Demszky et al. (2020) published a new paper where they introduce GoEmotion, a manually annotated dataset of 58 thousand Reddit Comments (in English), labeled for 27 emotions categories (based on semantic space theory).

The authors argue that in contrast to Ekman's taxonomy, which includes only a positive emotion (joy), this new taxonomy proposal includes a more significant number of positive, negative, and ambiguous emotion categories, which may be suitable for comprehension tasks that require an understanding of emotional expressions in conversations, such as analyzing feedback sent by a customer or interacting with chatbots.

To create this fine-grained dataset, the authors used a data dump that contains comments from 2005 to January 2019. They selected subreddits with more than 10k comments and applied a series of data curation (which is not specified in the original paper) to ensure the data does not reinforce language bias. They removed subreddits based on "not safe for work" public list. They kept vulgar comments because they could include negative emotions. They also used a pilot model (trained with 2.2K annotated examples) to exclude subreddits that consist of more than 30% of neutral comments or less than 20% negative, positive, or ambiguous comments. They also used a pilot model to balance the emotions in the dataset.

The annotation process consisted of three raters for each example. When no raters agree on at least one emotion label, they assigned two additional raters. All raters were native English speakers from India, but the authors did not detail their background (education, profession, and so on).

The authors used the dataset to train two models: a baseline model using a BiLSTM and a BERT-base (Devlin et al., 2018) for the final experiment. For the BERT model, they added a dense output layer on top of the pre-trained model for finetuning, with a sigmoid cross-entropy loss function to support multi-label classification.

They kept most of the hyperparameters presented by the original paper. They only change the batch size (to 16) and learning rate (to 5e-5). They also found that training for at least four epochs is indicated for this dataset, but training for more epochs results in overfitting.

The performance of the best model, BERT, on the test set, achieved an average F1-score of .46 (std=.19) for the full taxonomy (27 emotions). The BiLSTM model performed significantly worse than BERT, obtaining an average F1-score of .41 for the full taxonomy.

Authors suggest that it is a promising approach integrating affective science and NLU, but there is much room for improvement. In this sense, we investigated the performance of different transformers-based language with this dataset. A similar approach to the one driven by (Adoma et al., 2020), but using a fine-grained dataset of emotions

## 3 Experimental protocol

One of the many challenges of NLU is to identify the best embeddings to extract specific semantic information (for example, emotions) and identify different techniques to deal with long-term dependency. In recent years, we have seen different architectures emerging to deal with these challenges, such as RNNs, LSTM, BiLSTM. More recently, the Transformers architecture (Vaswani et al., 2017) has shown to be alternatives in addressing these limitations.

More recently, transformers' language models have shown better language understanding abilities to achieve state-of-the-art results in many different NLP and NLU tasks.

It is important to highlight that different implementations of language models have emerged from the original idea of the Transformer architecture. We hypothesize that different models can present different performance (f1-score) and time to complete in the task of recognizing emotions.

BERT was one of the first models based on Transformers to achieve great results in many NLP tasks. However, we assume that more recent Transformers language models could be more efficient than BERT. Thus, our goal is to investigate different transformer language models' behavior in this type of task.

### 3.1 Data

In this research, the dataset we will use is GoEmotion, released by Demszky et al. (2020). It is the largest manually annotated dataset of 58 thousand Reddit Comments (in English), labeled for 27 emotion categories (based on semantic space theory) or neutral.

Other annotated datasets are available for emotion recognition tasks, but they mainly rely on basic emotion theory from Paul Ekman, which comprises only six basic emotions. For this reason, we decided to use GoEmotion and investigate how different transformer language models can handle a dataset of fine-grained emotions.

### 3.2 Metrics

We will compute three standard metrics to measure the model's performance: precision, recall, and F1-score for each class. We will use F1-score (macro) as used in the original paper of GoEmotion (Demszky et al., 2020) to compare the performance among all the models.

We will also measure the time taken for each model to run to completion on training and evaluation.

### 3.3 Models

In the original paper, Demszky et al. (2020) used a BiLSTM as baseline model. The BiLSTM model performed achieved an average F1-score of 0.41. They also used a BERT implementation as the main model.

Google released the Bidirectional Encoder Representations from Transformers (BERT) pre-trained language model (Devlin et al., 2018) in 2018. It is considered the milestone of a rediscovery of the NLP area for improving performance in different natural language tasks. In the GoEmotion original paper, the BERT model performed significantly better than BiLSTM, with an average F1-score of 0.46 against an average F1-score of 0.41 from the baseline model.

In this research, we will assume BERT as our baseline model. Despite the excellent results in different benchmarks, this is a model that has some limitations. Since the release of BERT, different models were proposed to address some BERT limitations. For this reason, we will investigate the performance of four recent transformer language models (DistilBert, RoBERTa, XLNet, and ELECTRA) in the Emotion Recognition task. We will assume BERT as our baseline model.

DistilBERT is a distilled version of BERT, which is smaller, faster, cheaper, and lighter. This model is inspired by the Knowledge distillation approach. It is a compression technique to train a small model to reproduce a larger model's behavior (that is the reason it is also called teacher-student learning). Using this technique reduces the size of a BERT model by 40%, while 97% of its language capabilities are kept. The model is also 60% faster (Sanh et al., 2019).

RoBERTa (Robustly optimized BERT pre-training Approach) is a model developed by a Facebook team to improve BERT implementation. The researchers proposed some changes to the original model (Liu et al., 2019).

First, they used a larger dataset. BERT was trained on a combination of BookCopus plus English Wikipedia text, which totals 16GB of text. RoBERTa was trained with those two corpora plus three corpora from different domains: CC-News, OpenWeb Text, and Stories, which totals 160GB of text.

They also proposed improvements to the model design. For the training procedure, instead of using Next Sentence Prediction (NSP) task from BERT's pre-training, they introduced dynamic masking so that the masked token changes during the training epochs. They also trained the model on longer sequences (Liu et al., 2019).

XLNet (Yang et al., 2019) is a large bidirectional transformer that uses improved training methodology, larger data, and more computational power. To improve the training, XLNet developers introduced permutation language modeling. In contrast to BERT, which predicts only the masked 15% tokens (Masked tokens), XLNET predicts all tokens but in random order. XLNet was trained with over 130 GB of textual data. In addition to BERT's two datasets, they included three more corpora: Giga5, ClueWeb, and Common Crawl. XLNet outperformed BERT on 20 tasks, such as question answering, natural language inference, sentiment analysis, and so on.

ELECTRA is a model which training process does not rely on masked language such as BERT or RoBERTa. Instead of masking the input and predicting it, the ELECTRA approach replaces some tokens with plausible alternatives generated by a small generator network. Then, the model does not predict the original identities of the token replaced, and it trains a discriminative model to predict whether a generator sample replaced tokens in the input. The authors argued that this approach works well at scale. The model had the same performance as RoBERTa and XLNet but using less than ¼ of the computing resources. The model also outperformed them when the same computing amount was available (Clark et al., 2020).

### 3.4 Parameters Settings

When finetuning all Pre-trained transformers language model, we keep most of the hyperparameters set by Devlin et al. (2019) intact and only change the batch size and learning rate, based on the settings proposed by Demszky et al. (2020). We set the learning rate to 5e-5 and batch size to 16. We train the model for four epochs. The threshold to set a result as positive was 0.30. Parameter settings for all models are detailed:

- **Learning rate:** 5e-5
- **Batch Size:** 16
- **Epochs:** 4
- **Threshold:** 0.30

All models were implemented using the `huggingface` library. The training process used the same computing environment (Tesla T4 GPU).

### 3.5 Results

Table 4 summarizes our four models' performance (DistiBERT, RoBERTa, XLNet, and Electra) plus our BERT Baseline based on the original paper of GoEMotion (Demszky et al., 2020).

The model that achieved the highest f1-score (macro-average) was RoBERTa with .49 (std=.23). The model obtained the best result for 14 out of a total of 28 classes. The model did not achieve the worst performance for any class of emotions.

Two models achieved the second-highest f1-score (macro-average) but with different standard deviations. DistilBERT obtained the result of .48 (std = .21) while XLNet obtained the result of .48 (std = .23). The DistilBERT model achieved the best results for six classes of emotions (annoyance, gratitude, joy, realization, optimism, sadness) and the worst result for one class (love). T

XLNET model had the best result for nine classes of emotions (anger, approval, caring, confusion, curiosity, fear, gratitude, nervousness, remorse) and two worst results (amusement and pride). It is important to note that only the BERT and DistilBERT models achieved results for the pride class (.36 and .22, respectively), while the other models had an f1-score of zero.

ELECTRA was the worst model for the GoEmotion dataset. It achieved F1-score of .33 (std=.30). The model achieved the worst performance for 18 out of 28 classes. It is important to highlight the model had an F1-score of .0 for nine classes (caring, desire, disappointment, disgust, embarrassment, excitement, grief, nervousness, pride, realization, and relief).

However, along with BERT, ELECTRA achieved the best performance for the neutral class.

| Emotion | BERT | Distil BERT | RoBERTa | XLNet | Electra |
|---|---|---|---|---|---|
| admiration | 0.65 | 0.71 | 0.73 | 0.73 | 0.71 |
| amusement | 0.80 | 0.79 | 0.79 | 0.78 | 0.79 |
| anger | 0.47 | 0.49 | 0.48 | 0.51 | 0.47 |
| annoyance | 0.34 | 0.40 | 0.38 | 0.37 | 0.29 |
| approval | 0.36 | 0.37 | 0.38 | 0.38 | 0.31 |
| caring | 0.39 | 0.43 | 0.47 | 0.48 | 0.00 |
| confusion | 0.37 | 0.43 | 0.43 | 0.44 | 0.34 |
| curiosity | 0.54 | 0.55 | 0.55 | 0.57 | 0.53 |
| desire | 0.49 | 0.53 | 0.58 | 0.56 | 0.00 |
| disappointment | 0.28 | 0.24 | 0.35 | 0.32 | 0.00 |
| disapproval | 0.39 | 0.39 | 0.43 | 0.41 | 0.35 |
| disgust | 0.45 | 0.47 | 0.49 | 0.47 | 0.00 |
| embarrassment | 0.43 | 0.54 | 0.57 | 0.55 | 0.00 |
| excitement | 0.34 | 0.33 | 0.34 | 0.32 | 0.00 |
| fear | 0.60 | 0.62 | 0.68 | 0.68 | 0.37 |
| gratitude | 0.86 | 0.90 | 0.90 | 0.90 | 0.90 |
| grief | 0.00 | 0.00 | 0.00 | 0.00 | 0.00 |
| joy | 0.51 | 0.57 | 0.57 | 0.56 | 0.53 |
| love | 0.78 | 0.77 | 0.79 | 0.78 | 0.78 |
| nervousness | 0.35 | 0.34 | 0.34 | 0.35 | 0.00 |
| optimism | 0.51 | 0.59 | 0.59 | 0.58 | 0.54 |
| pride | 0.36 | 0.22 | 0.00 | 0.00 | 0.00 |
| realization | 0.21 | 0.28 | 0.26 | 0.28 | 0.00 |
| relief | 0.15 | 0.00 | 0.00 | 0.00 | 0.00 |
| remorse | 0.66 | 0.71 | 0.70 | 0.77 | 0.63 |
| sadness | 0.49 | 0.55 | 0.55 | 0.53 | 0.48 |
| surprise | 0.50 | 0.53 | 0.58 | 0.56 | 0.47 |
| neutral | 0.68 | 0.66 | 0.66 | 0.65 | 0.68 |
| **macro avg** | 0.46 | 0.48 | 0.49 | 0.48 | 0.33 |
| **std** | 0.19 | 0.21 | 0.23 | 0.23 | 0.30 |

Table 4: Results of different models for GoEmotion Taxonomy

Table 5 summarizes the time to complete training (4 epochs) and evaluation in the test set. The hyperparameters settings were the same for each model (learning rate and batch size), and they all ran in the same computing environment (Tesla T4 GPU).

| Time to complete | Training | Evaluation |
|---|---|---|
| BERT | 02:40:00 | 00:00:40 |
| DistilBERT | 00:33:58 | 00:00:09 |
| RoBERTa | 01:08:56 | 00:00:42 |
| XLNet | 01:33:49 | 00:01:06 |
| ELECTRA | 00:13:04 | 00:00:07 |

Table 5: Time to Complete for GoEmotion

XLNET was the model that took the longest to train, followed by RoBERTa, the model that achieved the best F1-score. ELECTRA was the fastest model of all, taking just 13 minutes in the training phase. However, it was the model with the worst performance. DistilBERT appears as an interesting model. It was a quick model to train (compared to the others), achieved the same F1-score as larger models (XLNet), and surpassed BERT.

## 4 Conclusion

Emotion Recognition in texts is an essential task in the field of affective computing. We investigated different language models' performance when using GoEmotion, a large manually annotated dataset for fine-grained emotion.

We used BERT as our baseline model and compared it with four additional transformers-based models (DistillBERT, RoBERTa, XLNet, and ELECTRA). Except for the ELECTRA model, which had the worst F1-score (.33), the other models had more similar results. RoBERTa achieved the best F1-score (.49), followed by DistillBERT (.48), XLNet (.48), and then BERT (.46). However, when we look at the metric of computational cost and time to complete, we argue that the DistillBERT model is the most efficient for this type of task.

The results show us that we still have much room to improve both models and create datasets for fine-grained emotions.